\documentclass[twocolumn,10pt]{asme2ej}

\usepackage{amssymb}
\usepackage{epstopdf}
\usepackage{etoolbox}
\usepackage{cite}
\usepackage{picinpar}
\usepackage{amsmath}               
  {
      \newtheorem{assumption}{Assumption}
       \newtheorem{remark}{Remark}
  }
\usepackage{url}
\usepackage{flushend}
\usepackage{textcomp}
\usepackage{bm}
\usepackage{comment}
\usepackage{makecell}
\UseRawInputEncoding
\usepackage{graphicx} 
\usepackage{epsfig}
\usepackage{fancyhdr}
\usepackage{setspace}
\usepackage{helvet}
\usepackage{hyperref}   
\hypersetup{
	colorlinks=true,
	linkcolor=blue,
	citecolor=blue,
	urlcolor=blue,
}
\usepackage[square,numbers]{natbib}

\title{Optimized Design of a Soft Actuator\\ Considering Force/Torque, Bendability, and Controllability via an Approximated Structure}

\author{Wu-Te Yang
    \affiliation{
	Mechanical Systems Control Laboratory\\
	Department of Mechanical Engineering\\
	University of California\\
	Berkeley, California 94720\\
        Email: wtyang@berkeley.edu
    }	
}

\author{Burak K{\"u}rk{\c{c}}{\"u}
    \affiliation{ 
        Assistant Professor\\
        Department of Computer Engineering\\
	Hacettepe University \\
        Ankara, Turkey 06800\\
        Email: bkurkcu@berkeley.edu
    }
}

\author{Masayoshi Tomizuka
    \affiliation{ 
        Distinguished Professor, Fellow of ASME\\
	Mechanical Systems Control Laboratory\\
	Department of Mechanical Engineering\\
	University of California\\
	Berkeley, California 94720\\
        Email: tomizuka@berkeley.edu
        }
}
\begin{document}
\maketitle
\begin{abstract}
{\color{blue}\it This paper introduces a novel design method that enhances the force/torque, bendability, and controllability of soft pneumatic actuators (SPAs). The complex structure of the soft actuator is simplified by approximating it as a cantilever beam. This allows us to derive approximated nonlinear kinematic models and a dynamical model, which is explored to understand the correlation between natural frequency and dimensional parameters of SPA. The design problem is then transformed into an optimization problem, using kinematic equations as the objective function and the dynamical equation as a constraint. By solving this optimization problem, the optimal dimensional parameters are determined. Six prototypes are manufactured to validate the proposed approach. The optimal actuator successfully generates the desired force/torque and bending angle, while its natural frequency remains within the constrained range. This work highlights the potential of using optimization formulation and approximated nonlinear models to boost the performance and dynamical properties of soft pneumatic actuators.}
\end{abstract}

\begin{nomenclature}
\entry{$A$}{cross-sectional area where the pressure p(y) is applied.}
\entry{$\textbf A$}{state space matrix of soft actuator's dynamic equation.}
\entry{$A_c$}{cross-sectional area of chamber of soft actuator.}
\entry{$A_s$}{inside cross-sectional area of syringe.}
\entry{$A_w$}{cross-sectional area of wall of soft actuator.}
\entry{$dA$}{small area where the pressure p(y) is applied.}
\entry{$dA_c$}{small area in the cross-section of a chamber.}
\entry{$dA_w$}{small area in the cross-section of the wall of soft actuator.}
\entry{$a$}{distance between the neutral surface and the top of soft actuator.}
\entry{\textbf B}{input matrix of soft actuator's dynamic equation.}
\entry{$b$}{distance between the neutral surface and the bottom of soft actuator.}
\entry{$C$}{damping constant of soft actuator.}
\entry{$\textbf C$}{output matrix of soft actuator's dynamic equation.}
\entry{$C_s$}{capacity of soft actuator.}
\entry{$E$}{Young’s modulus of material.}
\entry{$F(P)$}{equivalent force generated by pressure $P$.}
\entry{$F_m$}{measured force generated by pressure $P$ .}
\entry{$F_x$}{force inside the cantilever beam in the $x$ direction.}
\entry{$I_n$}{moment of inertia of soft actuator.}
\entry{$K$}{equivalent spring constant of soft actuator.}
\entry{$L_i$}{initial length of soft actuator.}
\entry{$L$}{elongated length of soft actuator.}
\entry{$\delta L$}{$L$ minus $L_i$.}
\entry{$M$}{mass of soft actuator.}
\entry{$P$}{input pressure generated by the syringe pump.}
\entry{$\dot P$}{derivative of input pressure generated by the syringe pump.}
\entry{$P_w$}{reaction pressure generated by the wall of soft actuator when P is applied.}
\entry{$p$}{scaling factor of Q matrix.}
\entry{$p(y)$}{pressure inside the beam and is a function of $y$.}
\entry{$Q$}{matrix to penalize states of soft actuator in LQR control.}
\entry{$Q_s$}{air output flow rate of the syringe.}
\entry{$R$}{radius of curvature when soft actuator is bending.}
\entry{$\textbf R$}{scalar to penalize input command in LQR control.}
\entry{$\mathbb{R}_+$}{set of positive real numbers.}
\entry{$T(P)$}{torque generated by soft actuator.}
\entry{$T_m$}{measured torque generated by soft actuator.}
\entry{$T_p$}{torque generated in soft actuator contributed by $P$.}
\entry{$T_{P_w}$}{torque generated in soft actuator created by $P_w$.}
\entry{$T_z$}{torque generated by pressure distribution function p(y).}
\entry{$V$}{Lyapunov function.}
\entry{$Y$}{Lyapunov matrix.}
\entry{$n$}{a parameter related to material properties.}
\entry{$t$}{wall thickness of soft actuator.}
\entry{$u$}{control input.}
\entry{$w$}{width of the actuator.}
\entry{$x$}{$x$ axis of coordinate system.}
\entry{$\textbf x$}{state vector includes $\theta$, $\dot\theta$, $\ddot\theta$.}
\entry{$y$}{$y$ axis of coordinate system.}
\entry{$z$}{$z$ axis of coordinate system.}
\entry{$\theta (P)$}{bending angle of soft actuator generated by input pressure $P$.}
\entry{$\dot\theta$}{velocity of bending angle of soft actuator.}
\entry{$\ddot\theta$}{acceleration of bending angle of soft actuator.}
\entry{$\zeta$}{damping ratio of soft actuator.}
\entry{$\Delta\zeta$}{perturbation of damping ratio of soft actuator.}
\entry{$\omega_m$}{motor speed of syringe pump.}
\entry{$\omega_n$}{natural frequency of soft actuator.}
\end{nomenclature}

\section{Introduction}

Soft robots have gained attention in recent years. Their degree of freedom, adaptability, and compliance are superior to the traditional robots. Soft robots show potential to explore unknown environments such as underwater or outer space explorations~\cite{c1, Tang2023unknown, trivedi2008optimal, c2}, deliver delicate components in medical industry~\cite{alici2018bending}, and manipulate fragile objects in the food industry~\cite{c4, dai2023soft, c35}. The motion of soft robots relies on soft actuators, and soft pneumatic actuators (SPAs) are the most popular options~\cite{c5, Tolley2018design, zuo2019design}. They are easier to fabricate, cost-effective, and have high power density~\cite{c31}. However, despite these advantages, the use of soft robots also presents challenges. Their elasticity reduces their generated force/torque and makes them hard to control. The task of optimizing force/torque, bending angle, and controllability of SPAs, which involves formulating an optimal design problem and defining suitable objective functions and constraints, poses a challenge to researchers~\cite{c9}.

In addressing these challenges, an intuitive design approach might draw inspiration from nature~\cite{c14, c15, Iida2023design}. Soft actuators inspired by natural forms, such as the human hand~\cite{dai2022design}, octopus arm~\cite{c16, c17}, or elephant trunk~\cite{Guan2020elephant}. Alternatively, finite element analysis (FEA) is commonly employed, along with optimization methods or machine learning algorithms, for exploring dimensional parameters of soft pneumatic actuators~\cite{c7,c8,c9,c10,Cecilia2018FEA}. While these existing approaches have been successful in creating effective soft actuators; however, they often focused on enhancing single performance metrics such as force/torque or bending angle. {\color{blue} Notably, the optimal design of SPAs considering multiple performance indexes is still seldom discussed. Additionally, those methods depend on trial-and-error and time-consuming simulations or experiments~\cite{c7}. This research, thus, aims to optimize force/torque and bendability, and improve controllability with model-based optimization formulation.}

Moreover, determining the dynamical properties of SPAs such as natural frequency is another challenge during the design stage.  As the optimal parameters are determined by optimization formulation, the dimensional parameters directly influence the natural frequency, which in turn impacts its controllability~\cite{c18, c31}.  Determining the dynamic properties, such as natural frequency, during the design phase is useful for enhancing the controllability of SPAs, influencing the system's pole locations, response time, and control efforts to achieve the desired response. However, modeling of those properties of SPAs is difficult due to their inherent compliance and nonlinearity~\cite{c41,c42}. Recent studies~\cite{c36, c37, c38} have addressed this challenge by simplifying SPA modeling, treating them as second-order systems where the natural frequency becomes a key parameter for improving controllability. 

\begin{figure}[t]
    \centering
    \includegraphics[width=175pt]{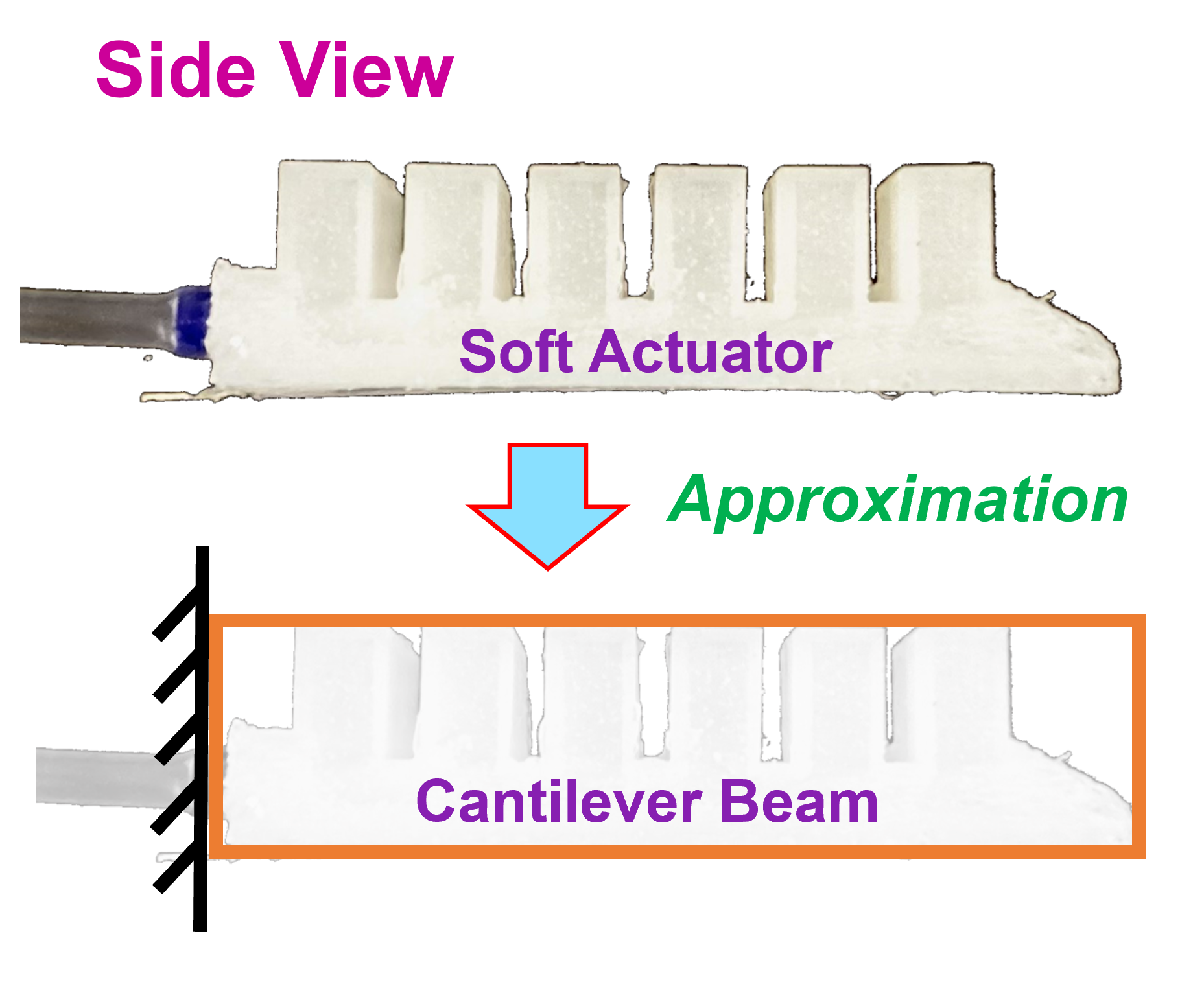}
    \caption{The soft pneumatic is analyzed mechanically by approximating its intricate structure as a cantilever beam.}
    \label{fig: 1}
\end{figure}

{\color{blue}Extending these insights, this paper introduces an optimal design approach for soft actuators considering multiple performance indexes.} To simplify the complex geometry of the soft actuator, it is approximated by a cantilever beam, as illustrated in Figure~\ref{fig: 1}. The optimization problem formulation links input air pressure to force/torque as well as bending angle, while also exploring the relationship between natural frequency and dimensional parameters. 
Preliminary tests and experiments validate our optimal design, showcasing enhanced output torque and bending angle compared to our previous work~\cite{c31}, while also modifying the dynamical properties. To the best of our knowledge, there is limited research that specifically examines and considers the dynamical properties of soft actuators during the design phase. The main contributions include:
{\color{blue}
\begin{enumerate}
    \item[$\bullet$] Deriving nonlinear kinematic and dynamic models based on approximated structure to facilitate a model-based optimal design formulation.
    \item[$\bullet$] Optimized design of a SPA, improving force/torque, bending angle, and system controllability concurrently, setting a new benchmark in multifaceted performance enhancement during the design phase.
    \item[$\bullet$] Validating the design and fabrication of SPA for controllability by linear quadratic optimal theory with achieved high-speed responses.
\end{enumerate}}

To position our contributions within the existing literature, we compare our approach with several design methods. Our previous work~\cite{c31} presented an optimal model-based design method to enhance the output force/torque of the soft actuator. However, we have modified the optimization formulation, which enhances force/torque, bending angle, and the system's controllability simultaneously. The bendability and natural frequency of the SPA are improved in this paper. Lotfiani et al.~\cite{c30} proposed a similar model-based optimal design method for a soft pneumatic actuator. However, the model was based on the hyperelastic model. In contrast, our approach uses nonlinear approximated models that are more implementable. Liu et al.~\cite{c29} introduced an energy-based method for searching optimal dimensional parameters, resulting in superior output torque compared to commercial soft actuators. Nonetheless, our method focuses on improving force/torque, bending angle, and system's controllability during the design phase. Demir et al.~\cite{c3} employed a machine learning algorithm to model pneumatic actuator performance using FEA simulation data. Their model was utilized to search for optimal design parameters under various constraints. In our work, we analyze and establish the mathematical models between input pressure and force/torque, bending angle, and natural frequency to identify the optimal parameters.
Polygerinos et al.~\cite{c19} attempted to correlate pressure changes and output torque in a soft pneumatic glove; however, their analysis relies on a mechanical model that requires the determination of multiple material properties through uniaxial tensile tests. In contrast, we approximate the structure to a simplified beam, reducing the number of parameters that need to be characterized. By simplifying the model, we trade complexity for accessibility.
In summary, this research aims to provide valuable insights to designers of a soft actuator design method that minimizes the need for trial-and-error methods. Meanwhile, the design approach enhances force/torque and bending angle and determines the system's dynamical properties that improve the controllability.

The remainder of this paper is organized as follows. Section 2 describes the kinematic and dynamic modeling of the SPA. Section 3 discusses the optimization formulation and optimal design of the SPA. Section 4 uses experimentation to verify the optimal design approach, and Section 5 concludes the work.

\section{System Modeling}
\label{sec:model}
{\color{blue}In this section, both nonlinear kinematic and dynamic models are constructed using an approximated structure, as shown in Fig.~\ref{fig: 1}. Using an approximated structure holds the merit of formulating a feasible and solvable optimization problem~\cite{c48}. Although hyperelastic models are more accurate~\cite{c30} than approximated models, the complexities of the models may present challenges in solving the optimization problem. Thus, the nonlinear models with approximated structures are utilized. The comparisons between predicted results and experimental results can be observed in Sec.~\ref{preVer} and Sec.~\ref{torque} and ~\ref{angle}.

\begin{figure}[http]
    \centering
    \includegraphics[width=195pt]{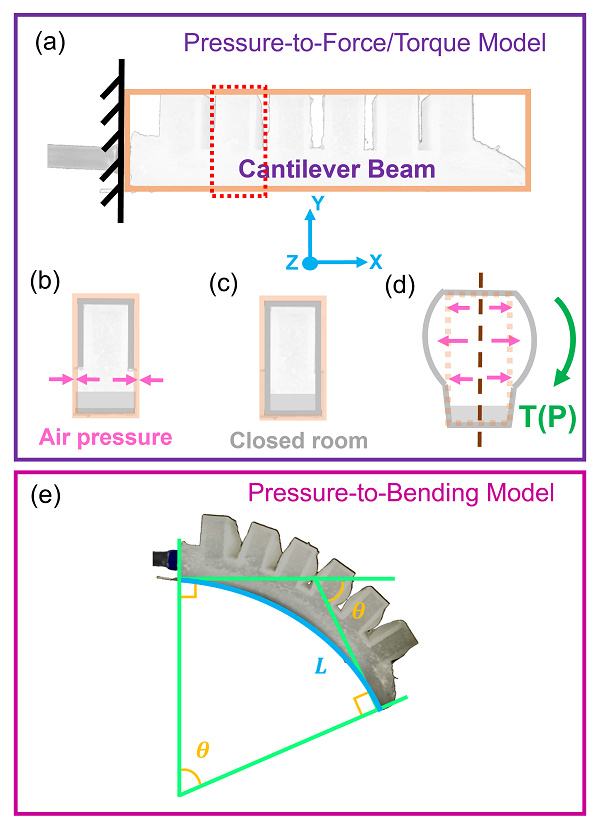}
    \caption{The mechanical analysis of the Pressure-to-Force/Torque model involves the following steps (a)-(d). (a) Segmenting a chamber from the actuator for analysis. (b) Both ends of the segmented chamber are open. (c) Treating the open chamber as a closed volume, as the air pressure within the chamber is evenly distributed and balanced across the open areas. (d) The supplied air pressure inflates the chamber and generates torques. The bending geometric of the soft actuator for analyzing the Pressure-to-Bending model is shown in (e).}
    \label{fig: 2}
\end{figure}

\subsection{Kinematic Modeling}
\label{sec:kinematic}
The soft pneumatic actuator features a corrugated geometric shape, characterized by a pattern of parallel ridges and grooves, and contains multiple discrete chambers. To analyze the structure mechanically, the corrugated structure can be approximated by the cantilever beam given by Fig.~\ref{fig: 1}. The approximation is grounded by the following assumption as in Sec.~\ref{optimization}.
Although some literature suggested linear model assumption\citep{c21,alici2018bending}, this work aims to consider a nonlinear model with an approximated structure.

\begin{assumption}\citep{c31}
\label{ass:uniform pressure}
The generated force/torque of the soft actuator is analyzed when the pressure distributes uniformly (steady state) across every individual chamber room.
\end{assumption}}

\begin{figure}[http]
    \centering
    \includegraphics[width=210pt]{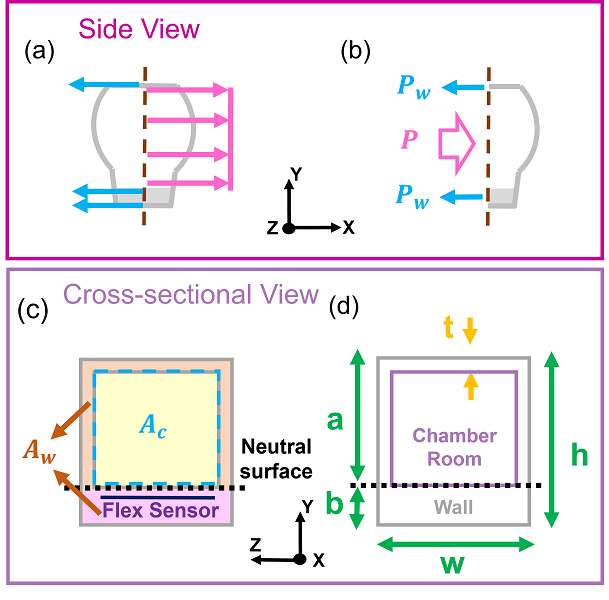}
    \caption{(a) The segmented chamber is halved to obtain the free-body diagram. (b) The free-body diagram showcases the pressures, $P$ and $P_w$, acting on it. (c) The cross-sectional view of the chamber is presented, where the neutral surface is depicted by a dashed black line and the embedded flex sensor is indicated by a purple line. (d) The dimensional parameters of the cross-section are established.}
    \label{fig: 4}
\end{figure}

The approximated structure is analyzed by the theories of mechanics~\cite{c18}. The obtained simplified models will serve as the objective function for the optimization problem in Sec.~\ref{optimization}.

\subsubsection{Pressure-to-Force/Torque Model}
The mechanical analysis process is demonstrated in Fig.~\ref{fig: 2} (a)-(d). A part of the cantilever beam (a chamber room) is segmented for mechanical analysis~\cite{c18}. The chamber is open on both sides because of the air channel. Although the chamber is open on both sides, it is treated as a closed chamber because of Assumption~\ref{ass:uniform pressure} and pressure balances in the open areas. Pressure applied to the chamber leads to structural expansion and generation of force/torque and bending angle. Given the continuity of the cantilever beam, the mechanical behaviors of the segmented parts are presumed to be continuous along the structure. Then, we use Eqn.~(\ref{eqn: 1}) to analyze the force equilibrium along {$x$} direction in Fig.~\ref{fig: 2}, and Eqn. (\ref{eqn: 2}) calculates the torque generated by the pressure supplied by the syringe pump~\cite{c32}:
\begin{align}
\sum{F_x} = \int_A{p(y)dA}
\label{eqn: 1}
\end{align}
\begin{align}
T = \sum{T_z} = \int_A{y \times p(y)dA}
\label{eqn: 2}
\end{align}
where $F_x$ is the force inside the cantilever beam in the $x$ direction, $p(y)$ is the pressure inside the beam and is a function of $y$, $T$ is the torque generated by the actuator, $T_z$ is the torque generated by pressure distribution function $p(y)$, and $dA$ is the small area where the pressure $p(y)$ is applied. Note that $p(y)$ becomes a constant under Assumptions~\ref{ass:uniform pressure}. 

The analysis of the force/torque generated by input pressure $P$ is presented in our previous work~\cite{c31}. The correlation between the input pressure $P$ and the pressure generated in the actuator's wall $P_w$ is given: 
\begin{align}
{P_w} = \frac{(a-t)(w-2t)}{bw+wt+2at-2t^2}{P}
\label{eqn: 3}
\end{align}
where $a$ is the distance between the neutral surface and the top of the actuator, $b$ is the distance between the neutral surface and the bottom of the actuator, $w$ is the width of the actuator, and $t$ is the wall thickness as shown in Fig.~\ref{fig: 4}(d). Equation (\ref{eqn: 2}) and (\ref{eqn: 3}) then produce a relationship between $P$ and generated torque $T(P)$. Therefore, the $T(P)$ can be computed as
\begin{align}
{T(P)} = {T_P} + {T_{P_w}}
\label{eqn: 4}
\end{align}
where $T_P$ is the torque contributed by $P$, and $T_{P_w}$ is the torque created by $P_w$ as shown in Fig.~\ref{fig: 4}(b). Since $P_w$ can be replaced by Eqn.~(\ref{eqn: 3}), $T_P$ and $T_{P_w}$ are described by using Eqn.~(\ref{eqn: 2}):
\begin{align}
{T_P} = \int_{A_c}{y \times Pd{A_c}}
\label{eqn: 5}
\end{align}
\begin{align}
{T_{P_w}} = \int_{A_w}{y \times \frac{(a-t)(w-2t)}{bw+wt+2at-2t^2}{P}d{A_w}}
\label{eqn: 6}
\end{align}
where $dA_c$ is the arbitrary small area in the cross-section of a chamber (yellow area) in Fig.~\ref{fig: 4}(c), $dA_w$ is the small area in the cross-section of the wall of the actuator (light orange and pink areas) in Fig.~\ref{fig: 4}(c) and $y$ is the location where the pressure acts as Fig.~\ref{fig: 4}(b). The material above the neutral surface is assumed to be in tension, producing positive internal pressures.

\subsubsection{Pressure-to-Bending Model}

{\color{blue}When it comes to the design of soft actuators, bending angle and force/torque are usually discussed in tandem. The bending angle is another index to evaluate the performance of soft actuators. Thus, enhancing the bending angle is another design objective.

The Pressure-to-Bending model will be built by referencing Euler's bending theory~\cite{c18, alici2018bending}. The Pressure-to-Bending model is inspired by the work, but there are some differences such as the geometric shapes of SPA. Also, the torque in our model is computed by the Pressure-to-Force/Torque model, while the torque in the work~\cite{alici2018bending} is calculated by a geometric method. Last but not least, the nonlinear bending model for large deflection components is considered~\cite{large2002Lee} where the corresponding bending theory is given by
\begin{align}
    \begin{split}
    \theta(P) = (\frac{n}{n+1})({\frac{T(P)}{EI_n}})^{\frac{1}{n}}L\\
    \label{eqn_ba1}
    \end{split}
\end{align}
\begin{align}
    \begin{split}
    I_n = (\frac{1}{2})^{(1+n)}(\frac{1}{2+n})w(a+b)^{(2+n)}
    \label{eqn_ba2}
    \end{split}
\end{align}
where $n \geq 1$ varies with soft materials and obtained through fitting the true $\sigma - \epsilon$ data provided by~\cite{Marechal2021material}, $\theta$ is the bending angle and is a function of $P$, $E$ is Young's modulus, $I_n$ is the moment of inertia for large deflection component, and $T(P)$ is obtained by (\ref{eqn: 4}). If considering the deformed length of the structure, we have
\begin{align}
\theta(P) = (\frac{n}{n+1})({\frac{T(P)}{EI_n}})^{\frac{1}{n}}(L_i + \delta L) 
\label{eqn_ba2}
\end{align}
where $L_i$ is the initial length, $L$ is the elongated length, and $\delta L$ = $L - L_i$. Since $\delta L$ = $PA_c{L_i}/{A_w}E$ in~\cite{c18}, Eqn.~(\ref{eqn_ba2}) becomes
\begin{align}
\theta(P) = (\frac{n}{n+1})({\frac{T(P)}{EI_n}})^{\frac{1}{n}}{L_i}(1+\frac{PA_c}{A_wE})
\label{eqn_ba3}
\end{align}
As $T(P)$ includes $P$, the $\delta L$ is also a function of $P$, bending angle is a function of $P$ and $P^2$. Both Pressure-to-Bending and Pressure-to-Force/Torque models serve as the objective function in Sec.~\ref{optimization}.}

\begin{figure}[http]
    \centering
    \includegraphics[width=190pt]{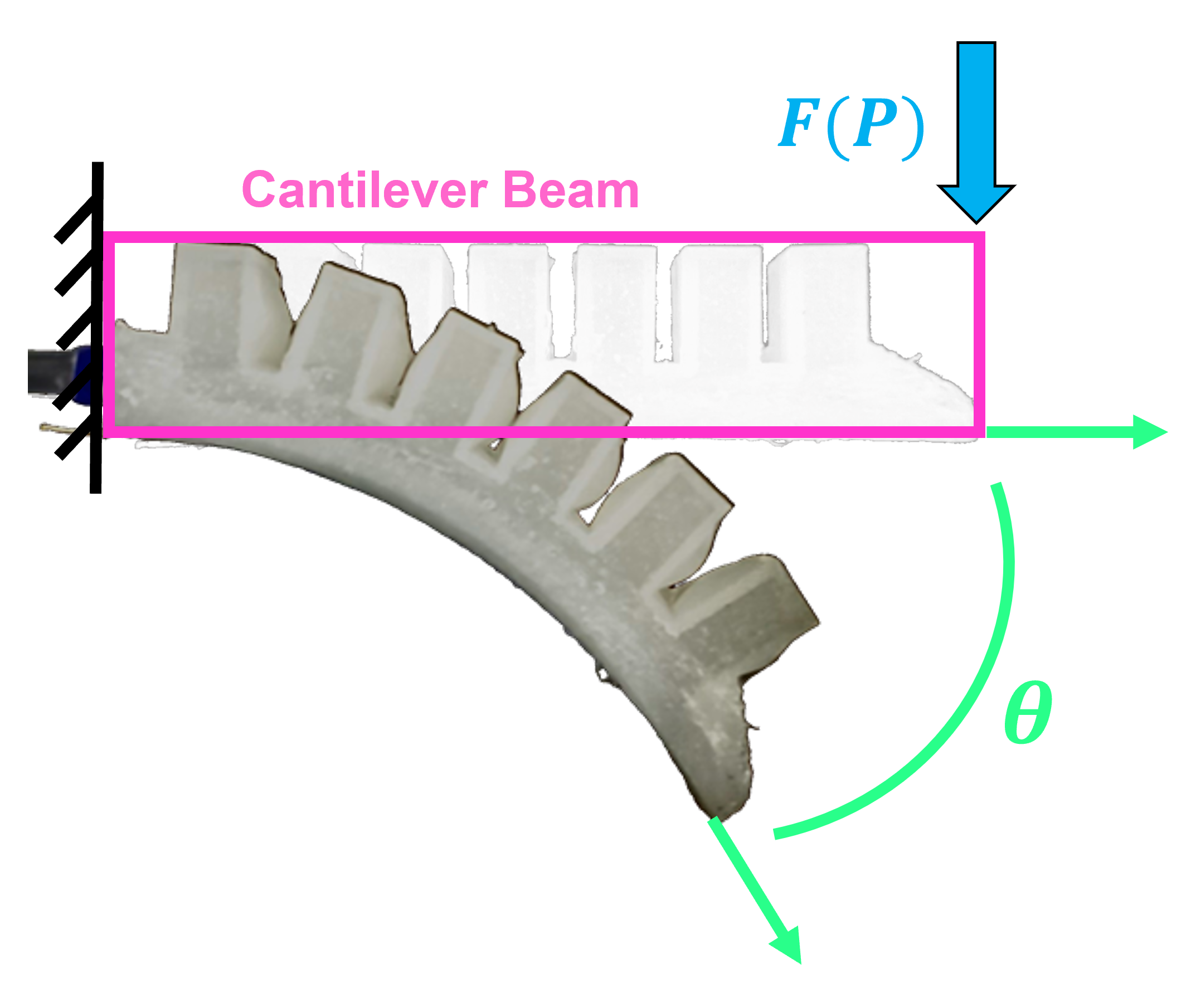}
    \caption{The approximated structure of the soft actuator generates a bending angle θ with a load F(P).}
    \label{fig: 5}
\end{figure}

\subsection{Dynamical Modeling}

{\color{blue}A second-order nonlinear model~\cite{large2002Lee} is employed to represent the dynamics of the bending angle in the soft actuator. This model is chosen for its mathematical simplicity and its capability to capture key dynamic properties: inertia, damping, and stiffness. Note that neglected dynamics in the modeling are considered as uncertainties, which can be compensated for in the control design phase~\cite{c36, c37, c38, c39}. The second-order dynamical model with nonlinear spring can be represented in the form as:
\begin{align}
\ddot{\theta} + \frac{C}{M}\dot{\theta} + \frac{K}{M}{\theta}^n = \frac{F}{M}
\label{eqn: 7}
\end{align}
Here, $M$ corresponds to the mass of the soft actuator, $C$ represents the damping constant of the system, $K$ and $F$ denote the spring constant of the system, and the force applied to the beam due to the input pressure, respectively. The values $\frac{C}{M}$ and $\frac{K}{M}$ can also be expressed as $2{\zeta}{\omega_n}$ and ${\omega_n}^2$, respectively, where ${\zeta}$ represents the damping ratio and ${\omega_n}$ denotes the natural frequency. Hence, the equation can be rewritten in a compact form as:
\begin{align}
\ddot{\theta} + 2{\zeta}{\omega_n}\dot{\theta} + {\omega_n}^2{\theta}^n = {F}/{M}
\label{eqn: 8}
\end{align}

The analysis can be visualized in Fig.~\ref{fig: 5}. When a force resulting from input pressure is applied at the end of the beam, the beam structure undergoes deflection and exhibits a bending angle $\theta$. The static equilibrium bending angle can be described by the Eqn.~(\ref{eqn: 9})~\cite{c18, large2002Lee}:
\begin{align}
\theta^n = (\frac{n}{n+1})^n \frac{FL_i^{n+1}}{EI_n}
\label{eqn: 9}
\end{align}
Here, $\theta$ represents the bending angle of the approximated beam under large deflection, and $F$ is the force applied at the end of the beam due to the input pressure. By manipulating Eqn.~(\ref{eqn: 9}), we obtain:
\begin{align}
\frac{F}{\theta ^n} = (\frac{n+1}{n})^n\frac{EI_n}{{L_i}^{n+1}} = {K}
\label{eqn: 10}
\end{align}
In Eqn.~(\ref{eqn: 10}), $K$ represents the equivalent spring constant of the approximated beam structure under the bending force. Therefore, we obtain the spring constant for Eqn.~(\ref{eqn: 7}). For simplicity, we temporarily neglect the damping term. Thus, Equation~(\ref{eqn: 7}) can be written as:
\begin{align}
\ddot{\theta} + (\frac{n+1}{n})^n\frac{EI_n}{{ML_i}^{n+1}}{\theta}^n = \frac{F}{M}
\label{eqn: 11}
\end{align}
The term $(\frac{n+1}{n})^n\frac{EI_n}{{ML_i}^{n+1}}$ in Eqn.~(\ref{eqn: 11}) represents the square of the natural frequency, as shown in Eqn.~(\ref{eqn: 8}). The natural frequency is given by:
\begin{align}
\omega_n = \sqrt{(\frac{n+1}{n})^n\frac{EI_n}{{ML_i}^{n+1}}}
\label{eqn: 12}
\end{align}}
\begin{remark}
The natural frequency of the system has a direct impact on its controllability. Higher values accelerate response times but might lead to increased energy consumption and other practical constraints. Conversely, lower natural frequencies could pose control challenges due to slower responses. Therefore, selecting a natural frequency requires a careful balance between responsiveness, energy efficiency, and system constraints. Ultimately, while increased natural frequency can enhance controllability, it is vital to consider its implications on system requirements and limitations. It is important to note that considerations such as excitation and the proximity of frequencies necessitate the evaluation of natural modes for each design. In this study, we only focus on the first natural frequency, as the second one appears at a considerable distance (three times farther) where the digital controller does not excite this mode.   
\end{remark}

{\color{blue}To complete the dynamical model (\ref{eqn: 8}), the damping ratio is estimated by second-order system identification over the step responses of the SPA. The damping ratio is not a constant but is accompanied by a perturbation term, ${\zeta}+{\Delta\zeta} = 0.7\pm0.1$, due to the nonlinearity and unpredictability of the soft materials. The completed dynamical model is needed and used for controller design in Sec.~\ref{SPAControl}}. 

\section{Optimal Design Analysis}
\label{optimization}
{\color{blue}When it comes to soft actuator design, optimal design for the force/torque does not imply optimal design for the bending angle. Usually, an optimal designed soft actuator on the bending may not generate large force/torque, and vice versa. This research tries to balance two performance indexes in the optimization formulation and design an optimal soft actuator.}
\subsection{Optimization Formulation}
In this subsection, our objective is to identify the optimal dimensional parameters of SPA. By utilizing the derived models in Sec.~\ref{sec:model}, we transform the design problem into an optimization problem based on the models, as presented by~\cite{c22}. We consider the mathematical models, $T(P)$ (\ref{eqn: 4}) and $\theta(P)$ (\ref{eqn_ba3}) as the objective function, subject to dimension constraints involving $a$, $b$, $w$, and $t$. Here, $a$ and $b$ represent the height of the chamber's cross-section, $w$ denotes the width, and $t$ signifies the wall thickness of the soft actuator. Figure~\ref{fig: 4}(d) presents a schematic representation of the dimensional parameters within the actuator's cross-section. {\color{blue}Additionally, since $T(P)$ and $\theta(P)$ are in different units, both equations should be normalized to put equal weight on both indexes. The $T(P)$ is normalized by 0.4 $Nm$ and becomes $\Bar{T}(P)$, while $\theta(P)$ is normalized by $1.4\pi ~rads$ and becomes $\Bar{\theta}(P)$. If different normalization factors are selected, different optimal parameters will be obtained.} The optimization problem is defined as
\begin{equation}
    \begin{split}
    \max_{a,b,w,t} & {~\Bar{T}(P) + \Bar{\theta}(P)} \\
    \textrm{s.t.} &~\dot P = 0\\
    &{a_1} \leq a \leq {a_2}    \\
    &{b_1} \leq b \leq {b_2}    \\
    &{h_1} \leq a+b \leq {h_2}    \\
    &{w_1} \leq w \leq {w_2}    \\
    &{t_1} \leq t \leq {t_2}    \\
    \label{eqn: 13}
    \end{split}
\end{equation}
where $P$ is a constant value and the constraint parameters $a$, $b$, $w$ and $t$ are determined by referencing human fingers’ dimensions~\cite{c23, c29}. {\color{blue}The parameters vary as follows: a ranges from 2 to 5 mm, b from 14 to 24 mm, w from 10 to 30 mm, t from 1.5 to 3.0 mm, and h from 15 to 25 mm. The similar results of Eqn.~(\ref{eqn: 13}) can be referenced in~\cite{c31}.}

\subsection{Considering System Controllability}
\label{control}
{\color{blue}Based on Eqn.~(\ref{eqn: 12}), the natural frequency of the soft pneumatic actuator is influenced by the dimensional parameters (moment of inertia). Equation~(\ref{eqn: 13}) can be revised by adding an additional constraint. Therefore, not only the optimal dimensional parameters will be identified but also the suitable natural frequency can be determined in the design stage. By squaring Eqn.~(\ref{eqn: 12}), we have
\begin{align}
{\omega_n}^2 = (\frac{n+1}{n})^n\frac{EI_n}{{ML_i}^{n+1}}
\label{eqn: 14}
\end{align}
and $I_n = (\frac{1}{2})^{(n+1)}(\frac{1}{n+2})w(a+b)^{(n+2)}$. Thus, Eqn.~(\ref{eqn: 14}) can be manipulated as
\begin{align}
{Ew(a+b)^{n+2}} = (\frac{n}{n+1})^n(2^{n+1}){(n+2)M{\omega_n}^2 {L_i}^{n+1}}
\label{eqn: 15}
\end{align}
Since the $L_i$ is constant, the designer can choose the ideal range of the natural frequency. 
\begin{equation}
    \begin{split}
    \max_{a,b,w,t} & {~\Bar{T}(P) + \Bar{\theta}(P)} \\
    \textrm{s.t.} &~\dot P = 0\\
    &{a_1} \leq a \leq {a_2}    \\
    &{b_1} \leq b \leq {b_2}    \\
    &{h_1} \leq a+b \leq {h_2}    \\
    &{w_1} \leq w \leq {w_2}    \\
    &{t_1} \leq t \leq {t_2}    \\
    &{C_1} \leq {Ew(a+b)^{n+2}} \leq {C_2}    \\
    \label{eqn: 16}
    \end{split}
\end{equation}
where the additional constraint of Eqn.~(\ref{eqn: 16}) with respect to  Eqn.~(\ref{eqn: 13}) is
\begin{equation}
C_1 \leq Ew(a+b)^{n+2} \leq C_2
\label{eqn: 17}
\end{equation}
Given that $E$ is a constant, determined by the selected material, and $w, a, b$ are bounded as in Eqn.~(\ref{eqn_opt1}), there are $a_3, a_4, b_3, b_4, w_3,$ and, $w_4$ which are inside the ranges of Eqn.~(\ref{eqn_opt1}), and we can say the minimum value for $Ew(a+b)^{n+2}$ is $Ew_3(a_3+b_3)^{n+2}$ and the maximum value is $Ew_4(a_4+b_4)^{n+2}$. 

\begin{equation} 
\begin{split}
    a_1 &\leq a \leq a_2\\
    b_1 &\leq b \leq b_2\\
    w_1 &\leq w \leq w_2\
\end{split}
\label{eqn_opt1}
\end{equation}

To ensure the added constraint does not render the problem infeasible, we need to choose $C_1$ and $C_2$ such that: 

\begin{equation}
\begin{split}
    C_1 = (\frac{n}{n+1})^n(2^{n+1}){(n+2)M_1{\omega_{n1}}^2 {L_i}^{n+1}}\\
    C_2 = (\frac{n}{n+1})^n(2^{n+1}){(n+2)M_2{\omega_{n2}}^2 {L_i}^{n+1}}
\end{split}
\label{c1c2}
\end{equation}
where $M_1$ and $M_2$ are the lower and upper bounds of mass of the SPA. Since mass is related to dimensional parameters of $a, b, w$ and $t$ (volume $\times$ density), the selection of the mass should reference the ranges of dimensional parameters in Eqn.~(\ref{eqn: 16}) to avoid hitting their upper or lower bounds. 

\vspace{-0.1in}
\begin{table}[http]
\centering
\caption{\label{tab:Table 1}Optimal parameters and its variances}
\scalebox{0.9}{
    \begin{tabular}{|c c c c c|} 
    \hline
    & b [mm] & a [mm] & w [mm] & t [mm] \\ [0.5ex] 
    \hline
    Opt. parameters & 4.0 & 20.0 & 30.0 & 1.5 \\ [0.5ex] 
    Variance 1 & 4.0 & 20.0 & 29.0 & 1.5 \\[0.5ex] 
    Variance 2 & 4.0 & 20.0 & 28.0 & 1.5 \\[0.5ex] 
    Variance 3 & 3.5 & 20.5 & 30.0 & 1.5 \\[0.5ex] 
    Variance 4 & 3.0 & 21.0 & 30.0 & 1.5 \\[0.5ex] 
    Variance 5 & 4.0 & 20.0 & 30.0 & 1.75\\[0.5ex] 
    Variance 6 & 4.0 & 20.0 & 30.0 & 2.0\\[0.5ex] 
    \hline
    \end{tabular}}
    \vspace{-0.1in}
\end{table}

Noting that $E > 0$ and all of $w, a, b$, and $n$ are non-negative, the function $Ew(a+b)^{n+2}$ is increasing with respect to $w, a$, and $b$. Therefore, the bounds for $C_1$ and $C_2$ are valid.
This ensures that there exists a range of values of $a, b, w$, and $t$ such that the additional constraint and all the original constraints can be satisfied simultaneously.}

\begin{figure}[http]
    \centering
    \includegraphics[width=210pt]{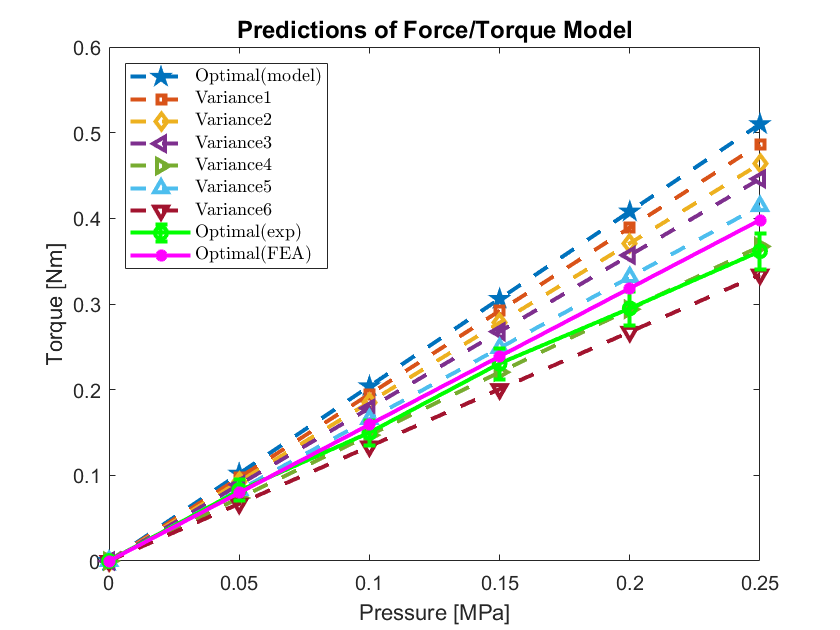}
    \caption{{\color{blue}Premilinarily verification of force/torque of optimal dimensional parameters by using the Pressure-to-Force/Torque model and the finite element analysis(FEA). The results are compared with the experimentation, Optimal(exp).}}
    \label{fig: 6}
\end{figure}

{\color{blue} The optimization problem~(\ref{eqn: 16}) was solved using the solver {\tt{fmincon}}, where the objective function and constraints were defined. The solver used interior-point algorithm~\cite{Byrd2000opt} to search for the optimal solution, employing a searching step size of $4 \times 10^{-12}$. The solution, which represents a minimum and satisfies all imposed constraints, is obtained after 42 iterations. Additionally, the sequential quadratic programming algorithm~\cite{numbericalOptimization} in {\tt{fmincon}} was also applied and the same parameters were obtained. Specifically, different initial value sets were tested and the optimization algorithm still obtained nearly the same optimal parameters.  The optimal values for the design parameters $a$, $b$, $w$, and $t$ are presented in Table~\ref{tab:Table 1}. Note that the parameters have been rounded to integers for manufacturability.} 

\begin{remark}
{\color{blue} The mass is not explicitly considered in Eqn.~(\ref{eqn: 16}), as it is indirectly influenced by dimensional parameters ($a, b, w$, and $t$). The mass is calculated as the product of the volume and density of the soft material. With the density being a constant determined by the chosen material, the volume of the soft actuator is solely dependent on $a, b, w$, and $t$. In addition, the constraints outlined in Eqn.~(\ref{c1c2}) dictate the allowable ranges ($M_1$ and $M_2$) of the SPA volume, thereby influencing the mass ranges.}
\end{remark}

\begin{figure}[http]
    \centering
    \includegraphics[width=210pt]{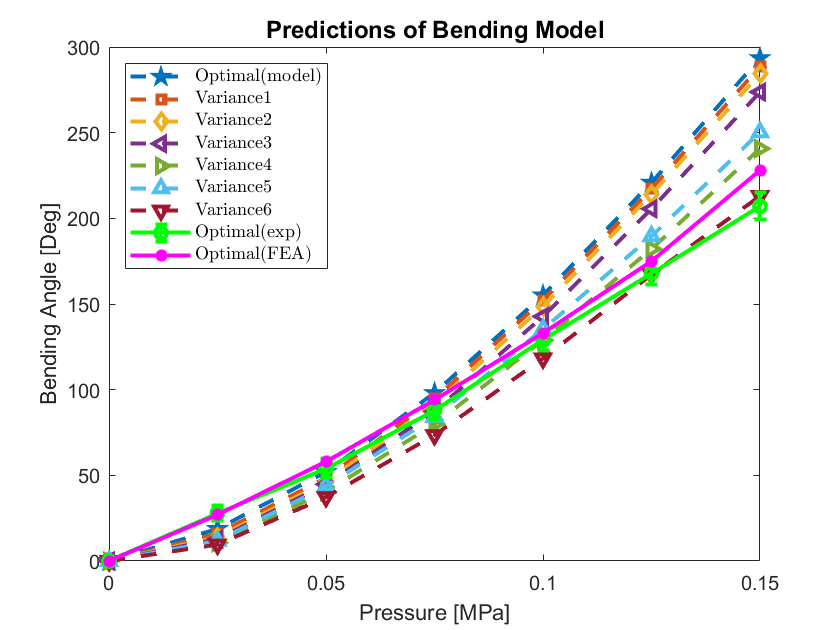}
    \caption{\color{blue}Preliminarily verification of bending angle of optimal dimensional parameters by using the Pressure-to-Bending model and the FEA. The results are compared with the experimentation, Optimal(exp).}
    \label{fig: 7}
    \vspace{-0.1in}
\end{figure}
\section{Experimental Evaluations}
\label{exp}

This section presents the analysis of both simulation and experimental results. To expedite the preliminary verification, predictions of kinematic models are implemented, allowing us to avoid the fabrication of all soft actuators listed in Table~\ref{tab:Table 1}. We investigate the impact of Eqn.~(\ref{eqn: 17}) on the optimal parameters and controllability of the system. Additionally, we present the results of essential performance metrics, namely torque and bending angle. Lastly, an LQR controller is designed to achieve desired dynamical performance, focusing on reducing response time and minimization of steady-state error.

\subsection{Preliminary Verification}
\label{preVer}
This section aims to first evaluate the performance of optimal design and its variances in Table~\ref{tab:Table 1} and then to avoid time-consuming fabrication when researchers design a soft pneumatic actuator. The popular pre-test tool for soft actuator design is the finite element method(FEA)~\cite{c7,Cecilia2018FEA}. However, this research derives kinematic models of soft pneumatic actuators in Sec.~\ref{sec:kinematic}. As the models are used to optimize the force/torque and bending angle of the soft actuator in Sec.~\ref{optimization}, the models are supposed to generate optimal force/torque and bending angle with optimal parameters. That is, the combination of $\Bar{T}(P)$ and $\Bar{\theta}(P)$ of optimal parameters, the objective function in Eqn.~(\ref{eqn: 16}), is supposed to be maximum. 

{\color{blue}Note that soft materials are set as an isotropic linear elastic in the FEA formulation.
The elastic modulus and Poisson’s ratio of the material, Ecoflex® Dragon skin 20, are 0.34 MPa and 0.49~\cite{c21}. The mesh size is approximately 1 mm.}

\begin{figure}[http]
    \centering
    \includegraphics[width=215pt]{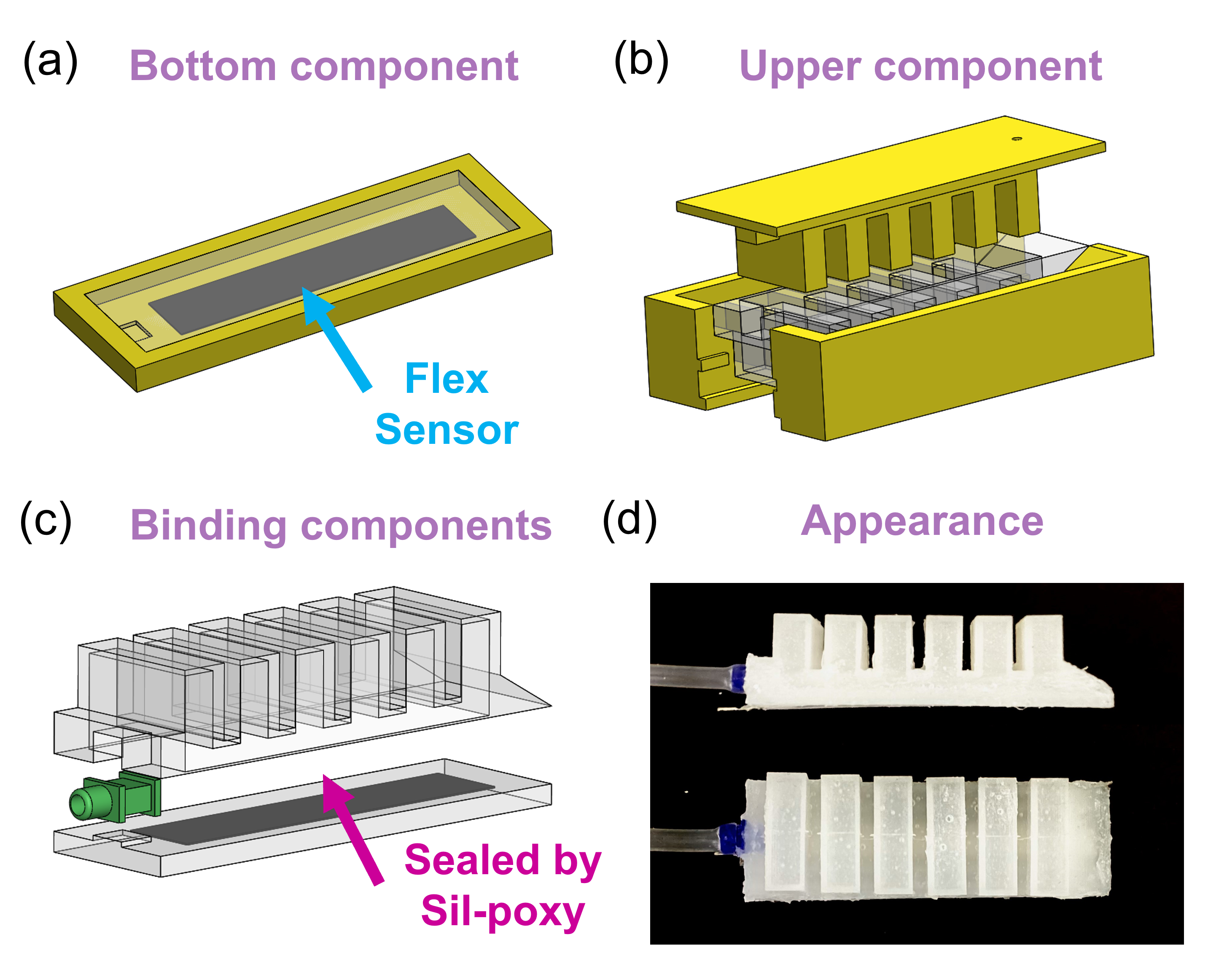}
    \caption{Both (a) and (b) illustrate the separate manufacturing of the bottom and upper components using distinct molds. (c) The soft actuator is demonstrated in an exploded view. The visual representation of the soft actuator's appearance can be seen in (d).}
    \label{fig: 3}
\end{figure}

\subsubsection{Preliminary Verification of Force/Torque}
{\color{blue}The Pressure-to-Force/Torque model can evaluate the torque of soft actuators with input pressures. The predicted, FEA, and experimentation results are displayed in Fig.~\ref{fig: 6}. Optimal design {\color{blue}(blue solid line)} outperforms all variances in both low and high pressures, generating up to 0.5 $Nm$ at 0.25 $MPa$. While the optimal design is superior, 'Variance 1' closely approaches its performance. Consequently, both optimal design and 'Variance 1' are selected to be fabricated and tested by experimentation.}

\subsubsection{Preliminary Verification of Bending Angle}
The Pressure-to-Bending model is applied to predict the bending angles of soft actuators with input pressures. {\color{blue}The predicted, FEA, and experimentation results are displayed in Fig.~\ref{fig: 7}}. Similarly, optimal design {\color{blue}(the same as in Fig.~\ref{fig: 6}) outperforms all variances in low and high pressures. Optimal design generates up to 290 $deg$ at 0.15 $MPa$ predicted by model.} Consequently, the bending angle preliminary test also verifies the optimal design.

Based on the results of both models, the optimal parameter set {\color{blue}(Design 1)} generates a maximum combination of $\Bar{T}(P)$ and  $\Bar{\theta}(P)$, {\color{blue}which is 2.36 at 0.15 $MPa$. Because the results of Variance 1 (2.30 at 0.15 $MPa$)} and optimal parameters are closed, both parameter sets in Table~\ref{tab:Table 1} are selected for manufacturing. Subsequently, those prototypes are utilized for experimental validation in Sec.~\ref{torque} and~\ref{angle}.

\begin{figure}[http]
    \centering
    \includegraphics[width=205pt]{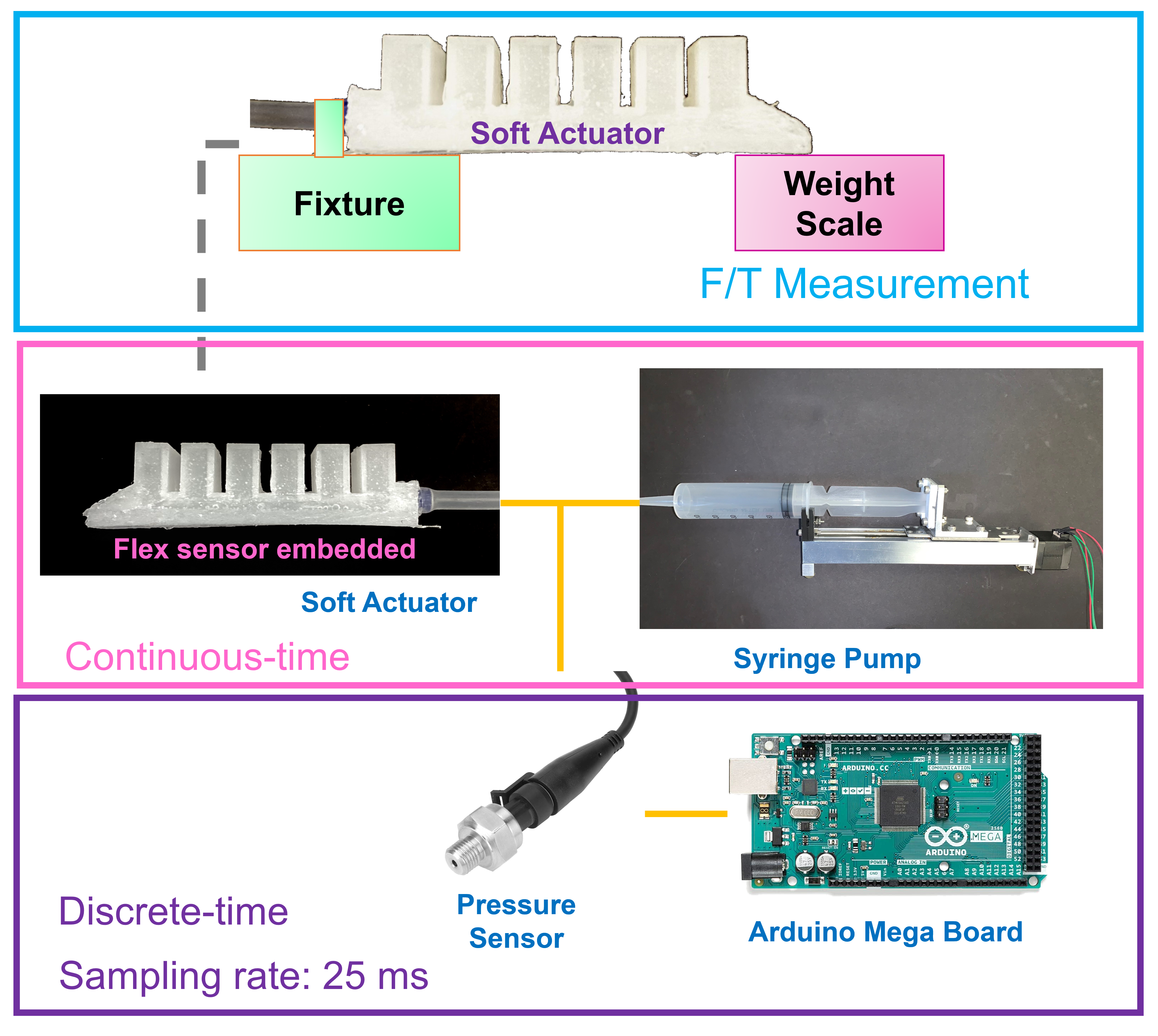}
    \caption{The schematic of the experimental setup.}
    \label{fig: 8}
    \vspace{-0.1in}
\end{figure}

\subsection{Study on Optimizing System's Controllability}
\label{controllable}

In Sec.~\ref{control}, in particular Eqn.~(\ref{eqn: 14})-(\ref{eqn: 15}), we know that the dimensional parameters will affect the natural frequency of the soft actuator. Thus, the optimization formulation is modified as Eqn.~(\ref{eqn: 16}). This subsection aims to explore the impact of the dimensional parameters on the system's controllability. Generally, the multiplication of the damping ratio and natural frequency affects the pole location and response time of the system based on Eqn.~(\ref{eqn: 8}). Since the damping ratio of the selected material is approximately 0.7, we mainly study the natural frequency ranging from 2 to 3.5 $rad/s$. Thus, the step responses of the designed SPA could be less than 2 to 3 $sec$ by experimental tests. Four different constraint ranges of Eqn.~(\ref{eqn: 17}) are selected in Table 2. The interior-point algorithm~\cite{Byrd2000opt} is used to search the optimal parameters with different constraint ranges of Eqn.~(\ref{eqn: 17}).

\begin{table}[http]
\centering
\caption{\label{tab:Table 2}The constraint ranges of natural frequency and the predicted natural frequencies are compared}
\scalebox{0.85}{
    \begin{tabular}{|c|c|c||c|c|c|c|} 
    \hline
    $\omega_{n1}$ & $\omega_{n2}$ & Real $\omega_{n}$ & b& a& w& t\\ [0.5ex] 
    [rad/s] & [rad/s] & [rad/s] &[mm] &[mm] &[mm] &[mm]\\ [0.5ex]
    \hline
    2.50 & 3.50 & 2.86 & 4.0 & 20.0 & 30.0 & 1.5 \\
    2.40 & 2.60 & 2.49 & 4.0 & 19.3 & 30.0 & 1.5 \\
    2.20 & 2.40 & 2.26 & 4.0 & 16.8 & 30.0 & 1.5 \\
    1.60 & 1.80 & 1.83 & 4.0 & 14.4 & 30.0 & 1.5 \\
    [0.8ex] 
    \hline
    \end{tabular}}
\end{table}

\begin{figure}[http]
    \centering
    \includegraphics[width=210pt]{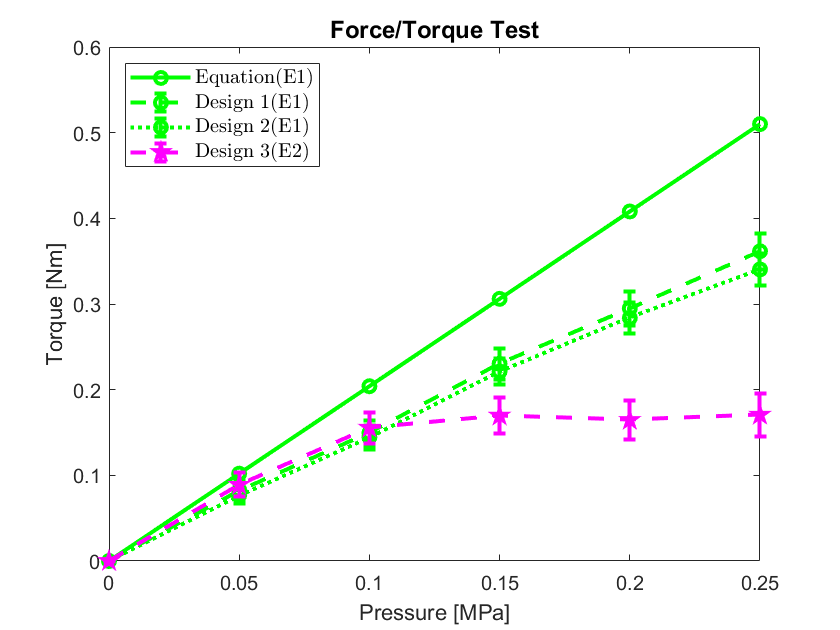}
    \caption{The experimental validation of optimal designed soft actuator, Design 1(E1). Compared to Design 1(E1), Design 2(E1) has a slightly smaller width. A softer material is used in Design 3(E2) and other dimensional parameters are the same.  E1 represents Dragon Skin 20, while E2 depicts Dragon Skin FX-Pro.  }
    \label{fig: 9}
    \vspace{-0.1in}
\end{figure}

{\color{blue}Various constraint ranges of Eqn.~(\ref{eqn: 17}) are displayed in Table~\ref{tab:Table 2}.  The corresponding optimal parameters are also listed in Table~\ref{tab:Table 2}. Different constraint ranges of the natural frequency yield different sets of optimal parameters. The height of the soft actuator, which directly influences the moment of inertia, varies with the range of the natural frequency. Thus, the moment of inertia will influence the natural frequency as Eqn.~(\ref{eqn: 14}). The real natural frequency of the last prototype in Table~\ref{tab:Table 2} hit its upper limit. It is caused by applying an approximated structure in Eqn.~(\ref{eqn: 16}). The experimental results are demonstrated in Sec.~\ref{NF and Control}. Note that real natural frequencies in Table~\ref{tab:Table 2} are obtained by system identification the experimental step responses of those soft actuators.}

\subsection{Fabrication of Soft Actuator}
The configuration of the proposed soft actuator is given by Fig.~\ref{fig: 3}(c). The actuator's body is primarily made of liquid rubber, specifically Ecoflex\textregistered Dragon Skin 20 and Ecoflex\textregistered Dragon Skin FX-Pro. The upper and bottom components shown in Fig.~\ref{fig: 3}(c) are fabricated using two distinct molds, as illustrated in Fig.~\ref{fig: 3}(a) and (b), respectively, inspired by~\cite{c25}. 

These components are then securely bonded together using the silicone adhesive, Sil-poxy\textregistered. The nozzle at the end is connected to a rubber tube, enabling the input of air from the syringe pump~\cite{c32} into the chambers. Notably, the bottom component incorporates a flex sensor, inspired by~\cite{c19, c43}, which is embedded inside to regulate the position of the neutral surface. In Fig.~\ref{fig: 4}(c), the thickness of the thin silicone layer between the flex sensor (indicated by the purple line) and the desired neutral surface (represented by the black dashed line) is precisely controlled during fabrication, ensuring a thickness of 0.5 $mm$. The resulting soft pneumatic actuator is presented in Fig.~\ref{fig: 3}(d), with dimensions measuring 24 $mm$ in height, 30 $mm$ in width, and 94 $mm$ in length.
\begin{figure}[http]
    \centering
    \includegraphics[width=210pt]{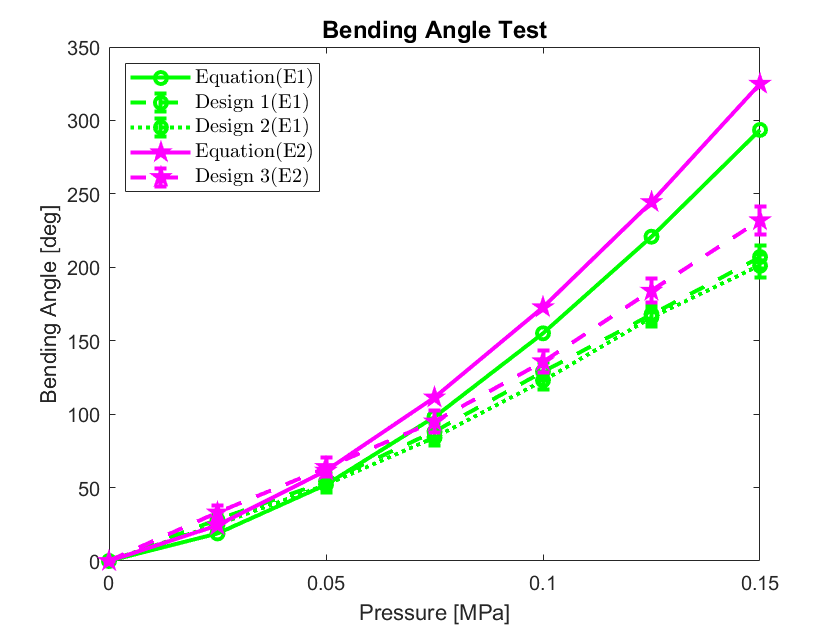}
    \caption{The test results for the bending angles of all designs are presented. 'Equation (E1)' predicts the bending angles of the soft material used in the production of Design 1(E1) and Design 2(E1). 'Equation (E2)' simulates the bending angles of the soft material employed in the fabrication of Design 3(E2).}
    \label{fig: 10}
\end{figure}

\subsection{Experimental Setup}
The experimental setup is illustrated in Fig.~\ref{fig: 8}. The soft actuator is powered by a self-designed syringe pump, which is driven by a stepper motor. The pressure is regulated by the position of the stepper motor.  Moreover, the embedded flex sensor(SparkFun Electronics, Niwot, CO) is used to monitor the bending angle of the soft actuator. To monitor the air pressure, a pressure sensor (Walfront, Lewes, DE) with a sensing range of 0 to 80 $psi$ is utilized, which is synchronized with an Arduino MEGA 2560 (SparkFun Electronics, Niwot, CO). The microcontroller is based on the Microchip ATmega 2560 with a sampling time of 25 $ms$. Furthermore, serial communication between the microcontroller and a computer is established by a USB cable.

\subsection{Test of Torque}
\label{torque}
One of the objectives of the proposed design approach is to optimize the output torque of the soft actuator. 
The fabricated soft actuator is fixed at the end of its structure as in the top of Fig.~\ref{fig: 8}. The tip of the soft actuator is contacted on a digital weight scale (Etekcity, Anaheim, CA) with a resolution of 0.01 $g$ and a range of 5 $kg$. As the air pressure is pumped into the actuator, the actuator inflates and exerts a force on the weight scale. The torque is computed by using the equation below
\begin{align}
{T_m} = F_m \times {L_i}
\label{eqn: 30}
\end{align}
where $T_m$ is the measured torque
, $F_m$ is the measured force, 
and $L_i$ is the length of the soft actuator.
The results in Fig.~\ref{fig: 9} show that the optimized design ('Design 1(E1)'), made of Dragon Skin 20, matches the trend of model predictions ('Equation(E1)'). The maximum torque of the optimal design is 0.359 $Nm$ at the pressure of 0.25 $MPa$. By comparison, the previous version~\cite{c31} has torque of 0.144 $Nm$ at the same pressure. Note that E1 represents the Young's modulus of Dragon Skin 20, while E2 depicts that of Dragon Skin FX-Pro. 

Once again, we conducted tests on alternative designs as an initial test of their suboptimality, specifically referring to 'Design 2(E1)' and 'Design 3(E2)'. In Design 2(E1) or Variance 1 in Table~\ref{tab:Table 1}, the width of the soft actuator is slightly smaller, 29 $mm$, in comparison to Design 1(E1). The performance of Design 2(E2) drops a little, but its performance is close to Design 1(E1). In the case of Design 3(E2), Young's modulus is lower, measuring 0.26 $MPa$, in comparison to Design 1(E1). Due to its softer nature, Design 3(E2) experiences buckling under higher loads, resulting in a torque plateau of approximately 0.15 $Nm$. It is worth noting that all designs exhibit similar performance up to 0.1 $MPa$ of pressure, indicating that the linear model assumption is most beneficial for applications under limited pressures.

\subsection{Test of Bending Angle}
\label{angle}
The proposed approach also enhances the bendability of soft actuators. Also, the bending angle serves as another important indicator to evaluate the performance of a soft actuator~\cite{c8,c9}. The flexibility of the soft actuator has been shown to have an impact on its output torque; when the actuator exhibits a larger bending angle under a fixed input pressure, it has the potential to generate higher torques~\cite{c8,c9}. To test the bending angle, we adopted a defined approach illustrated in Fig.~\ref{fig: 2} (e), where we mark the positions of the actuator's tip and end on a piece of grid paper. This method enables us to accurately quantify the bending angle of the soft actuator. 

As depicted in Fig.~\ref{fig: 10}, the results of the bending angle test reveal that Design 2(E1) exhibits a smaller degree of bendability. Design 3(E2) achieves a maximum bending angle of approximately 232 degrees, whereas Design 1(E1) and Design 2(E1) reach 206 and 201 degrees, respectively. The 'Equation(E1)' represents the predicted results of the Pressure-to-Bending model, taking into account Young's modulus of Ecoflex\textregistered Dragon Skin 20. This curve is compared with Designs 1(E1) and 2(E1), as they share the same material. In contrast, 'Equation(E2)' depicts the predicted outcomes of the model, considering the material properties of Ecoflex\textregistered Dragon Skin-FX Pro, and it is compared with Design 3(E2).

{\color{blue}Compared to the experimental results, the predictions of the Pressure-to-Bending model underestimate low bending angles and overestimate high bending angles. However, the model still catches the trend of the bending angle of soft actuators, which is beneficial for optimization formulation. Notably, the combination of real $\Bar{T}(P)$ and $\Bar{\theta}(P)$ of Design 1(E1) surpasses that of Design 2(E1) and Design 3(E2). The combined values of $\Bar{T}(P)$ and  $\Bar{\theta}(P)$ of each prototype are 1.44, 1.29, and 1.40 at 0.15 $MPa$, respectively. Therefore, experimental results validate the optimal design.}


\subsection{Verification of Natural Frequency and Pole Location}
\label{NF and Control}

The relationship between dimensional parameters and the natural frequency is discussed in Sec.~\ref{control} and \ref{controllable}. Firstly, we aim to verify the accuracy of Eqn.~(\ref{eqn: 12}). If the error of the equation is minimal, it can serve as a reference. The error ranges from approximately 5.76 \% to 16.86 \% as the Table~\ref{tab:Table 3}, which is influenced by the dimensional parameters and Young's modulus. For Design 1(E1) to Design 2(E1), they have the same shape, height, and wall thickness but have different widths. Design 2(E1) has the smallest error 5.76 \%. The softer the materials, the larger the errors of Eqn.~(\ref{eqn: 12}). For example, Design 3(E2) has an error of 16.86 \% because its structure will buckle.

\begin{table}[http] 
\setlength{\tabcolsep}{0.8pt}
\centering
\caption{\label{tab:Table 3}Comparisons of true natural frequencies and the estimations by the Eqn.~(\ref{eqn: 12})}
\scalebox{0.85}{
\begin{tabular}{|c c c c|}
\hline
Unit [rad/s] & True $\omega_n$ & Estimated $\omega_n$ & Error\\ [0.5ex]
\hline
\makecell{Design 1(E1)\\(\footnotesize E=0.34MPa, M=0.35N, L=0.94m)} & 2.86$\pm0.052$  & 2.62 & 8.39\%\\ [0.5ex] 
\hline
\makecell{Design 2(E1)\\(\footnotesize E=0.34MPa, M=0.34N, L=0.94m)} & 2.78$\pm0.048$ & 2.62 &  5.76\%\\ [0.5ex]
\hline
\makecell{Design 3(E2)\\(\footnotesize E=0.26MPa, M=0.46N, L=0.94m)} & 1.72$\pm0.061$ & 2.01 & 16.86\% \\ [0.5ex] 
\hline
\end{tabular}}
\end{table}

\begin{figure}[http]
    \centering
    \includegraphics[width=200pt]{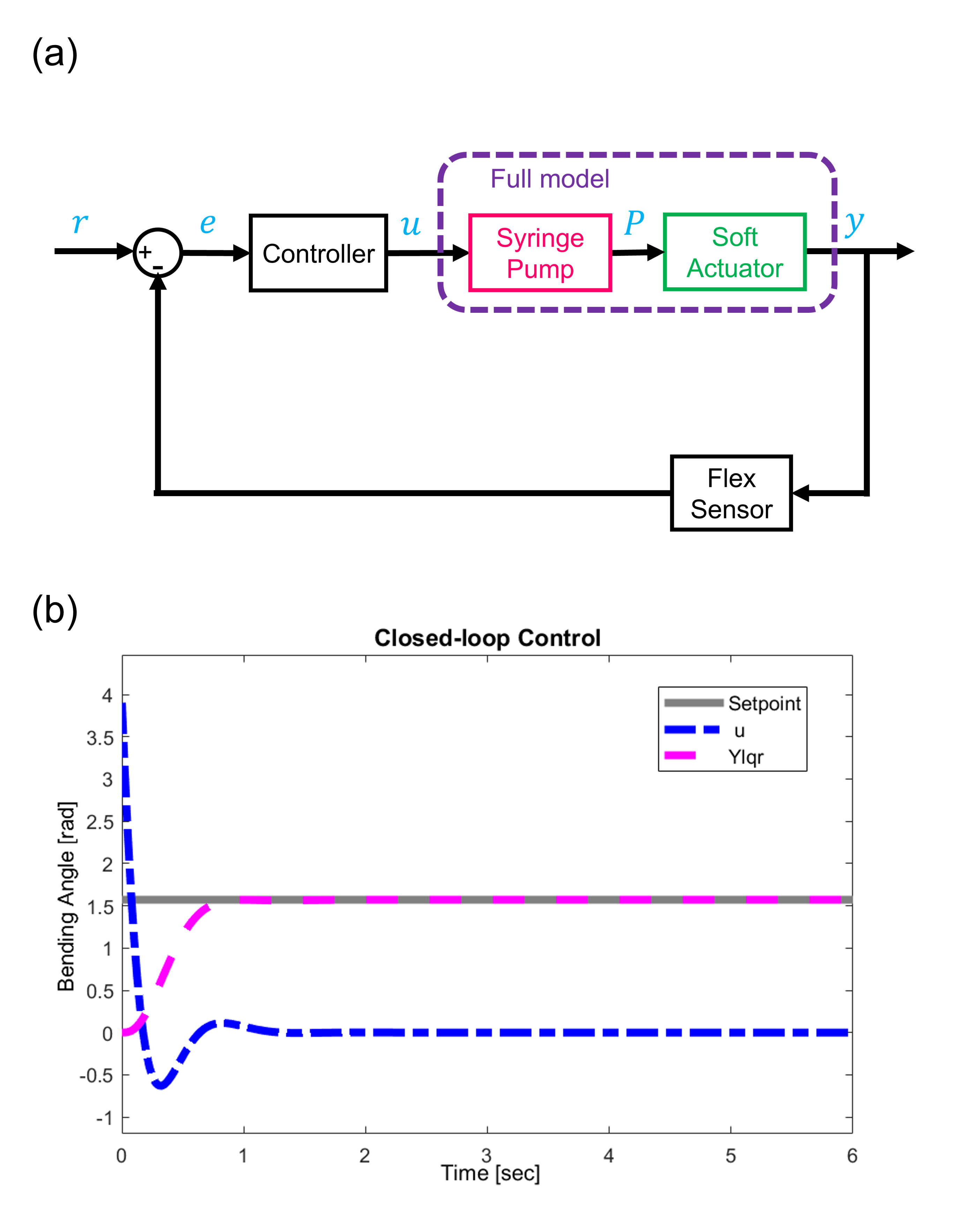}
    \caption{\color{blue}(a) The control block diagram of real experiment is visualized. (b) The pink dashed line displayed the step response of the soft actuator(Design 1(E1)) controlled by the LQR controller, the blue dashed line showed the control commands($u$) of LQR from the microcontroller in Fig. 8, and the gray solid line represented the reference.}
    \label{fig: 11}
\end{figure}
Then, the experiments are conducted to validate if the natural frequency lies within the desired range. In Table~\ref{tab:Table 2}, the optimal parameters are obtained with a constraint range of natural frequency between 2.5 and 3.5 $rad/s$. The corresponding pole location of the SPA lies between -2.45 and -1.75. The real natural frequency is measured by a second-order system identification over the responses of the soft actuator. The real natural frequency is 2.86 $rad/s$ as in Table~\ref{tab:Table 2}. The real locations of poles are -1.99 $\pm$ j2.04. The other prototypes in Table~\ref{tab:Table 2} are also verified by experimentation. The natural frequency lies either within or close to the desired range. Therefore, the results show that both the natural frequency and the pole location lie in the desired range. 

\subsection{Soft Actuator Control}
\label{SPAControl}
{\color{blue}Firstly, the Eqn.~(\ref{eqn: 8}) is linearized for controller design, which is valid even during a 90-degree bending}. The damping ratio of the soft actuator is 0.7 $\pm$ 0.1 obtained by second-order system identification. The natural frequency is estimated by Eqn.~(\ref{eqn: 12}). The $F$ in Eqn.~(\ref{eqn: 7}) is generated by input pressure $P$. Then, we have the dynamic model of the soft actuator with all the parameters in Eqn.~(\ref{eqn: 8}). In addition, the system is driven by a syringe pump which is a first-order system~\cite{c32}. The full model, thus, is a third-order system as below. 
\begin{align}
{T_{SYS}} = \frac{lA_s{\omega_m}c/2\pi{C_s}M}{s^3 + 2{\zeta}{\omega_n}s^2 + {\omega_n}^2s} 
\label{eqn: 18}
\end{align}
where ${Q_s}$ is the air output flow rate of the syringe, and $A_s$ is the inside cross-sectional area of the syringe, $C_s$ is a capacity of the soft actuator, ${\omega_m}$ is the motor speed, and $c$ is a constant obtained by experimentation.

Since the full system is known and is controllable, we employ a Linear Quadratic Regulator (LQR) for its advantageous ability to satisfy key performance specifications such as settling time and steady-state errors. The controller is implemented into a microcontroller, equipped with a saturation to account for the motor's speed limit as 5 $rev/s$.

We define the state vector as $\textbf{x}=[\theta~\dot{\theta}~ \ddot{\theta}]^T$. The full system model is then reformulated into a controllable canonical form as follows:
\begin{align}
{\bf \dot x} =  \textbf{A} {\bf x} + \textbf{B} {\bf u} 
\label{eqn: 19}
\end{align}
\begin{align}
{\bf y} = \textbf{C} {\bf x}
\label{eqn: 20}
\end{align}
with matrices defined as:
\begin{align}
\textbf{A} = \begin{bmatrix} 0 & 1 & 0\\0 & 0 & 1\\ 0 & -{\omega_n}^2 & -2{\zeta}{\omega_n} \end{bmatrix},
\textbf{B} = \begin{bmatrix} 0 \\0 \\ 1\end{bmatrix},
\textbf{C} =\begin{bmatrix} \frac{lA_s{\omega_m}c}{2\pi{C_s}M} \\ 0 \\ 0\end{bmatrix}^T
\label{eqn: 21}
\end{align}
With the directly measurable, via embedded flex sensor, $\theta$ and robust differentiation methods for approximating $\dot{\theta}$ and $\ddot{\theta}$~\cite{c47} with the sampling time of 25 $ms$, LQR control emerges as a viable and effective solution.

The LQR design aims to determine an optimal state-feedback controller, $u$, by minimizing a quadratic cost function of the form $J=\int_0^{\infty}({\bf x}^TQ{\bf x}+u^T\textbf{R}u)dt$. This objective function signifies the trade-off between striving to achieve desired state values and expending control effort. The scalar $\textbf{R}$ and the matrix $Q$ in the cost function are weighting factors that can be adjusted to prioritize control effort versus state deviations. In particular, the diagonal elements of $Q$ are chosen to penalize the deviation of states from their desired values. The off-diagonal elements of $Q$ would introduce coupling between the states in the cost function, but we keep them zero for simplicity.

For the system, $\textbf{R}$ is set to 1, and $Q$ is defined as follows:
\begin{align}
Q = p\begin{bmatrix} 1 & 0 & 0\\0 & 0.3 & 0\\ 0 & 0 & 0 \end{bmatrix}
\label{eqn: 23}
\end{align}
With $p \in \mathbb{R}_+$, this choice of $Q$ represents the independent treatment of each state. The highest weight is assigned to the bending angle $\theta$, reflected by the first diagonal element. The speed of bending $\dot \theta$, although important, is considered less significant, as shown by the second element set to 0.3. The third diagonal element set to 0 indicates we don't penalize changes in the acceleration $\ddot \theta$. The scalar $p = 100$ allows for global adjustment of state deviation tolerance against control effort. Depending on the system requirements, $p$ can be increased or decreased. This decision depends on whether we want to penalize the state deviations heavily (larger $p$) or keep the control effort low (smaller $p$). The choice of $Q$ should be customized based on the system's unique needs and may require iterative tuning for optimal performance.

Upon obtaining $Y$ by solving ${\textbf{A}^T}Y + Y{\textbf{A}} + Y\textbf{B}{\textbf{R}^{-1}}{\textbf{B}^T}Y + Q = 0$, we have the state-feedback controller as follows:
\begin{align}
{u} = -\textbf{R}^{-1}{\textbf{B}^T}Y{\textbf x}
\label{eqn: 24}
\end{align}
The stability of the system is evaluated using a Lyapunov function:
\begin{align}
{V} = {\textbf x}^T{Y}{\textbf x}
\label{eqn: 25}
\end{align}
\begin{align}
\dot{V} = \dot{\textbf x}^T{Y}{\textbf x} + {\textbf x}^T{Y}\dot{\textbf x} = {\textbf x}^T({\textbf{A}^T{Y}+Y\textbf{A}}){\textbf x}
\label{eqn: 26}
\end{align}
Because $Y > 0$, the Lyapunov function $V$ is positive definite and decrescent, while $-\dot{V}$ is positive definite. Consequently, the system is uniformly stable according to the Lyapunov stability criterion~\cite{c46}.

{\color{blue}Figure~\ref{fig: 11} (a) and (b) depict the block diagram and experimental step response of the Design 1(E1), respectively.} The LQR controller exhibits prompt and accurate response characteristics, with a settling time of approximately 0.8 $sec$ and an almost negligible steady-state error. These characteristics validate the LQR controller's ability to handle the presence of noise and system delays, as observed in our experiments, ensuring that the experimental results align closely with simulation results.

\section{Discussion and Conclusion}

\subsection{Discussion}
\label{discuss}
The system parameter in Eqn.~(\ref{eqn: 8}) has a correlation with the dimensional parameters such as width, height, and wall thickness. The variations of the dimensional parameters will influence the natural frequency of the system. Thus, the constraint space of the optimization formulation of Eqn.~(\ref{eqn: 16}) is extended to place the natural frequency. The influence of the dimensional parameters is examined in Sec.~\ref{controllable}. If the constraint space changes, the dimensional parameter, especially the height, will vary accordingly. The natural frequency will be moved as well. The optimal designs in Table~\ref{tab:Table 2} barely excite the second natural frequency since it is far away from their first natural frequency. Overall, we can not only enhance the force/torque and bending angle of the soft actuator but also place the pole location of the system. This approach can determine the dynamical property in the design stage.

Since the natural frequency of the soft actuator varies with the dimensional parameters, the constraints of dimensional parameters should be changed with the range of the natural frequency. In Table~\ref{tab:Table 2}, the dimensional parameter $a$ exhibits a strong correlation with the natural frequency range, thereby establishing its crucial role in the actuator design. For instance, when the natural frequency range was defined as 3.0 to 3.5 $rad/s$, the actual natural frequency deviated, reaching approximately 2.9 $rad/s$. Thus it was attributed to $a$ being at the threshold of its defined limit, 20 $mm$. To address this discrepancy and align the natural frequency with the desired range, an adjustment was required in the parameter $a$. Consequently, by extending the upper limit of $a$ to 24 $mm$, we were successful in tuning the natural frequency to the desired value of 3.15 $rad/s$. This underscores the importance of the iterative adjustment of parameters in achieving optimal performance, particularly when it comes to satisfying the constraints related to the natural frequency of the actuator.

Besides, Young's modulus of soft materials affects the bending angle and natural frequency of soft actuators. According to Eqn.~(\ref{eqn_ba3}), Young's modulus should be small to optimize the bending ability of soft actuators. If the constraint of Young's modulus is added, it always hits the lower bound. Meanwhile, Young's modulus, influenced by $a$, $b$, and $w$, is supposed to be at a certain range to achieve the desired dynamical properties by Eqn.~(\ref{eqn: 17}). This will lead to no solution when the intersection of Eqn.~(\ref{eqn: 17}) and the constraint of Young's modulus is empty. In other words, the optimal Young's modulus cannot be found by the optimization algorithm. Another issue is that smaller Young's modulus will cause soft actuators to buckle as the Design 3(E2) in Fig.~\ref{fig: 9}. Therefore, Eqn.~(\ref{eqn: 16}) does not include the constraint of Young's modulus, which is suggested to be selected by designers.

Last but not least, the number of chambers has an influence on the force/torque and bending angle of soft actuators in different ways. Based on Eqn.~(\ref{eqn_ba3}), an increase in the number of chambers (longer ${L_i}$) enhances the bending ability of soft actuators. Even if the number of chambers (${L_i}$) does not affect the generated torque by referencing Eqn.~(\ref{eqn: 4}), an increased number of chambers will result in the buckling of soft actuators. Thus, the generated torque will reach a plateau above a specific pressure value, which is similar to Design 3(E2) in Fig.~\ref{fig: 9}. Since the number of chambers influences force/torque and bending angle differently, this research does not add the number of chambers as a constraint in Eqn.~(\ref{eqn: 16}). Instead, our designs maintain a fixed number of chambers and manage the length of soft actuators to be approximately 100 $mm$, so those soft actuators can avoid the buckling issue.

\subsection{Conclusion}
{\color{blue}The study introduces an innovative approach to optimize the design of soft pneumatic actuators, focusing on enhancing force/torque, bending angle and improving the system's controllability. A cantilever beam approximation is implemented to analyze the complex structure of SPAs. This approximation allows for the derivation of both nonlinear kinematic and dynamic models. The design problem is converted to an optimization problem, with the kinematic models serving as the objective function, and the dynamical model as a constraint. This approach leads to the determination of optimal dimensional parameters for SPAs. To validate the effectiveness of the proposed method, preliminary verification and several experiments are conducted. The optimal soft actuator demonstrates the ability to generate a torque of up to 0.359 $Nm$ and a bending angle of 206 $deg$, while its natural frequency falls within the desired range. The output force/torque and bending angle outperform that of our previous design. Lastly, an optimal controller is designed to control the system which achieves 0.8 $sec$ settling time and almost 0 steady-state error. The relationship between dimensional parameters and natural frequency has been studied and discussed. This optimal model-based design strategy presents a novel method to enhance multiple performance indexes of the soft pneumatic actuator.}

\bibliographystyle{asmems4}

\bibliography{asme2e}

\newpage
\listoffigures

\vskip 1 in
This section will be created when figures are included with a caption.

\newpage
\listoftables

\vskip 1 in
This section will be created when tables are included with a caption.

\end{document}